# An Autonomous Free Airspace En-route Controller using Deep Reinforcement Learning Techniques

Joris Mollinga
University of Amsterdam
Amsterdam, The Netherlands
jorismollinga@gmail.com

Herke van Hoof
University of Amsterdam
Amsterdam, The Netherlands

*Abstract*—Air traffic control is becoming a more and more complex task due to the increasing number of aircraft. Current air traffic control methods are not suitable for managing this increased traffic. Autonomous air traffic control is deemed a promising alternative. In this paper an air traffic control model is presented that guides an arbitrary number of aircraft across a three-dimensional, unstructured airspace while avoiding conflicts and collisions. This is done utilizing the power of graph based deep learning approaches. These approaches offer significant advantages over current approaches to this task, such as invariance to the input ordering of aircraft and the ability to easily cope with a varying number of aircraft. Results acquired using these approaches show that the air traffic control model performs well on realistic traffic densities; it is capable of managing the airspace by avoiding 100% of potential collisions and preventing 89.8% of potential conflicts.

## I. INTRODUCTION

Traditional Air Traffic Control (ATC) is becoming more and more complex due to increasing number of aircraft but also new forms of air traffic control are required to manage drones and other (electrical) airborne vehicles. Especially low altitude airspace occupied by relatively small aircraft like drones and helicopters is currently mostly unregulated while predictions are that the number of aircraft in this airspace will increase [1]. This increase is partially caused by numerous companies around the world working on various applications of airborne vehicles. Recent work [10, 13, 8, 18] outlining the future of automated airborne transportation states that the demand for autonomous air traffic control systems is high. It is observed that current air traffic control systems are not suitable for this task. Deep Learning (DL) or Deep Reinforcement Learning (DRL) techniques are a promising way of accomplishing this as they have proven to be applicable in control problems.

During the past decade, Reinforcement Learning (RL) has become a major field of study within Artificial Intelligence (AI). In RL an agent attempts to maximize a reward signal by performing actions in an environment. The agent is not told what action to take, but learns from the reward function, and should learn to balance immediate rewards vs. delayed rewards. Many breakthroughs have been achieved over the past years, primarily in the field of DRL. Two well-known examples are AlphaGo beating world champion Lee Sedol in Go [16] and DeepMind achieving super human scores in Atari games [12]. Also in other, more sophisticated games like Dota 2, DRL algorithms are capable of beating amateur teams [2].

The inspiration for this research also comes from a game, Air Control Lite[1], where the goal is to guide aircraft to their runway for landing, like an air traffic controller. The job of an air traffic controller is to prevent collisions of aircraft, safely and efficiently organize the flow of traffic and to provide support to pilots. Although traffic flow and efficiency are important factors, their primary goal is to guarantee safety of the aircraft. To accomplish this, air traffic controllers use traffic separation rules which ensure the distance between each pair of aircraft is above a minimum value all the time. These rules are also the core of Air Control Lite and while this games provides the player with a simplified version of reality, it inspired us to think about what AI can contribute to ATC.

In this paper DRL techniques are applied to design an air traffic control model that can perform the task of en-route controller. This air traffic controller manages an unstructured airspace: aircraft are not limited to flying between waypoints, they can fly to any point in the three-dimensional airspace.

This paper is structured as follows. In section 2 a few examples of previous research on autonomous air traffic control is presented. In section 3 the RL, graph and graph-based DL frameworks needed to understand this paper are presented. In section 4 the experiment is described, followed by section 5 which details the implementation. Results are presented in section 6 while section 7 concludes this paper.

---

[1]  Downloadable for Android here:
*https://play.google.com/store/apps/details?id=dk.logisoft.airc ontrol&hl=en_US*

## II. RELATED WORK

Research on autonomous air traffic control has been conducted for decades. This section presents a few well known examples of this research and provides examples of research relevant to this paper.

One of the most well-known techniques in the field of autonomous air traffic control is the conflict resolution algorithm designed by Erzberger [5]. It can automate the task of an en-route controller but it can also handle departing and arriving aircraft. It consists of various components for conflict detection, conflict resolution and simulation. This algorithm iteratively evaluates generated resolution trajectories until a viable solution is found. These trajectories are generated with expert knowledge of air traffic controllers and with insights gained from operation and analytical studies. When evaluated on large airspaces it shows excellent performance. This algorithm is designed to mimic what an air traffic controllers would do. Using data driven approaches allows for learning new heuristics which can be more efficient than the heuristics used by air traffic controllers. In this research data driven approaches are investigated and evaluated.

Next to the conflict resolution algorithm discussed above Erzberger also worked on other topics in the field of autonomous air traffic control. Together with Itoh he presents an algorithm that helps with scheduling arriving aircraft into the approach area in [6]. The goal of the scheduler is to assign aircraft a runway for landing and to provide aircraft with a schedule containing times to cross certain waypoints on the way to the runway in a way that minimizes delays. The scheduler uses a centralized planner for calculating these time-slots for passing certain way-points by using minimum separation distances between different types of aircraft. Generally speaking, aircraft are scheduled on a first-come, first-served basis but it is possible to change the order of landing aircraft. The algorithms presented in [6] by Erzberger and Itoh form the basis for the Center/TRACON Automation System, which consists of multiple decision support tools for managing arriving traffic in the United States.

With recent advances in the field of machine learning and reinforcement learning, data driven approaches applied to the task of air traffic controller are also interesting to investigate. These approaches are interesting to investigate because they can lead to new insights or heuristics regarding airspace management and collision avoidance. In the remainder of this section, data driven approaches for the task of air traffic controller are discussed.

In [3] Brittain and Wei present an Artificial Neural Network (ANN) for avoiding conflicts on intersecting and merging airways. In this approach, each aircraft is represented as an agent. Next to the agents own information (speed, acceleration, distance to goal), each agents state-space also encompasses information on the $N$-nearest aircraft. The action space is limited to the one-dimensional domain and defined by three actions: slow down, take no action, speed up. It is trained using an actor-critic algorithm. The aircraft is penalized if the separation requirement between two aircraft is broken. For intersecting and merging airways case studies near optimal results are reported.

However, Brittain and Wei only consider a few case studies which all need a different model. Generalizing a single model to multiple case studies is not part of their research. Taking the $N$-nearest aircraft is also a disadvantage. In this context the input ordering of the $N$-nearest aircraft is important. A small change in the location of aircraft may yield a permutation of the ordering of the $N$-nearest aircraft and thus the input to the ANN. The ANN should learn that these permutations represent nearly the same airspace, which might be a difficult task.

Brittain and Wei try to circumvent this problem by using Long Short Term Memory Networks (LSTMs) in [4]. However, LSTMs are still not truly input order invariant. Ideally, a type of ANN is desired which can deal with a variable input ordering of aircraft. Graph based methods offer this advantage and are investigated in this research. Next to that, expanding to the three-dimensional domain is also done in this research. This provides for more opportunities for collision avoidance and resembles reality better, where aircraft can navigate in a three-dimensional environment.

## III. BACKGROUND

### A. Reinforcement learning

Reinforcement learning is a field in machine learning that studies how agents interact with an environment. Next to supervised learning and unsupervised learning, it is one of the three main pillars of machine learning. In reinforcement learning, training labels do not need to be provided (unlike in supervised learning) but the agent learns from taking actions and receiving a reward from the environment for this action. The goal of the agent is to maximize a function of the cumulative reward. Usually the environment is represented as a Markov decision process.

Markov Decision Processes (MDP's) are a framework to formalize a sequential decision making process. An MDP consists of an *agent* (or multiple *agents*) in an *environment*. Each agent is in a specific *state* and can take *actions* in this environment. For each action, the agent receives a *reward* and a *transition function* describes the transition behavior from one state to the next after taking a certain action. This framework is shown in figure 1. Formally, an MDP consists of the following 4-tuple: (*S, A, R, T*). Here *S are* the set of states, *A* the set of actions, $R(s_t, s_{t+1}, a_t)$, the reward function that yields reward $r_t$ when transitioning from state $s_t$ to $s_{t+1}$ after action $a_t$ and $T(s_t, s_{t+1}, a_t) = P(s_{t+1}/s_t, a_t)$, or the probability of transitioning to state $s_{t+1}$ when taking action $a_t$ in state $s_t$, respectively.

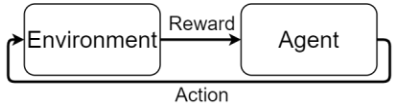

Figure 1. Reinforcement learning model

The process of taking an action, transitioning to the next state and receiving a reward is continuously repeated, creating a trajectory. The goal of the agent is to maximize some function of cumulative rewards over the trajectory, usually the discounted sum of rewards. This is given in (1):

$$G_t = \sum_{t=0}^{\infty} \gamma^t r_t \qquad (1)$$

In (1) $G_t$ is called the *discounted return*, $r_t$ is the reward at time step *t* and γ is the *discount factor* where 0<γ≤1, but usually it is close to 1. When γ<1, the return cannot become infinite, which could otherwise happen when the trajectory is infinite. If γ is close to 0, it values immediate rewards higher than rewards that accumulate in the long term. When γ is close to 1, it has a focus on maximizing the long term reward. To decide what to do, the agent tries to optimize its policy π, the strategy for taking action $a_t$ in state $s_t$. A policy can be deterministic (mapping *st* directly on to $a_t$) or stochastic ($s_t$ is mapped to a distribution of $a_t \in A$). How exactly a policy is learned is discussed in the next section.

### 1) Policy learning

If we consider a trajectory τ with discrete time steps τ =1...*T* and policy $\pi_\theta$ with parameters θ the goal of the agent is to maximize the expected return in this trajectory *G(τ)* as per (2).

$$J(\theta) = \mathbb{E}_{\pi_\theta}[G(\tau)] \qquad (2)$$

We can then then update θ using gradient ascent, using (3):

$$\theta_{t+1} = \theta_t + \alpha \nabla_\theta J(\theta) \qquad (3)$$

The expression for $\nabla_\theta J(\theta)$ differs per learning algorithm used. In this paper an actor critic model is used for training the air traffic control model. Actor-critic models consists an *actor* and a *critic*. The actor determines which action to take while the critic gives feedback on the actor, telling it how "good" this action was. For this algorithm $\nabla_\theta J(\theta)$ is given by (4) [17]:

$$\nabla_\theta J(\theta) = \mathbb{E}_\tau \left[ \sum_{t'=1}^{T} \nabla_\theta \log \pi_\theta(a_{t'}|s_{t'}) A(s_{t'}, a_{t'}) \right] \qquad (4)$$

Here $A(s_t, a_t)$ is called the *advantage-function*. This function can take on various forms but in this paper it is given by (5).

$$A(s_t, a_t) = \delta_t + (\gamma\lambda)\delta_{t+1} + \ldots + \ldots + (\gamma\lambda)^{T-t+1}\delta_{T-1} \qquad (5)$$
$$\text{where } \delta_t = r_t + \gamma V_v(s_{t+1}) - V_v(s_t)$$

A positive advantage function means that the actions that the agent took in the environment resulted in a better than average return, thus the probability of taking these actions is increased, and vice-versa if the advantage function is negative. This specific form of calculating the advantage function is called Generalized Advantage Estimation (GAE) [15]. λ is an additional hyper parameter allowing for a compromise between the bias and variance of the advantage estimate.

### 2) Reward shaping

An important part of implementing RL algorithms is the reward function. In practical applications however, reinforcement learning techniques suffer from the *sparse reward problem*. This means that rewards are so rare, that convergence of the algorithm is very slow and often intractable. One way to overcome this and to speed up learning is by defining a reward function that guides the agent towards the goal. This process is called *reward*. For example, an agent moving towards a goal can be given a positive reward if the distance to the goal is decreased.

Guiding the reward function in such a way can affect the policy that is learned. In fact, shaping functions can have adverse effects on the learned policy and after learning the agent can exhibit behavior that is not intended. To prevent these side effects, one needs a modified reward function that does not affect the optimal policy π*. Ng et al. present a framework for this in [14]. It introduces a specific type of shaping function called the potential function. They prove that the resulting policy from applying this function is consistent with the policy learned without shaping. The presented framework states that the reward function consists of two parts, which are added to get the reward at time step *t*. These parts are the unshaped reward signal $R(s_t, a_t, s_{t+1})$ and the shaping reward $F(s_t, a_t, s_{t+1})$. Ng et al. prove that if *F* is of the form presented in (6) then the optimal policy is unchanged by the shaping function. In (6) Φ is a potential function and can be any function that takes a state as input, but a clever implementation can greatly speed up learning.

$$F(s_t, a_t, s_{t+1}) = \gamma\Phi(s_{t+1}) - \Phi(s_t) \qquad (6)$$

### B. Graphs

Many different problems can be represented by nodes on a graph with dependencies and relations between nodes. For example social networks, or atoms in a molecule can be represented in a graph. For many different types of problems a graph representation is better suited for finding a solution to the problem. Graphs consist of a variable sized set of unordered nodes and edges. Formally, a graph is defined as the tuple (V, E), where V are the vertices or nodes of the graph and E the edges of the graph.

One way of representing $E$ is by using an *adjacency matrix*. The adjacency matrix $A$ is a square matrix of $N \times N$, which describes the connectivity of the graph. Here $N$ is the number of nodes in the graph. Typically $A[i, j] > 0$ if node $i$ and node $j$ are connected. The value at $A[i, j]$ can be the Euclidean distance between node $i$ and node $j$, or it can be a 1 if the nodes are connected and a 0 if they are not.

*C. Graph based DL techniques*

In this paper aircraft in an airspace are represented as nodes in a graph. Nodes in this graph are connected if the distance between a pair of aircraft is below a certain threshold. This approach offers a number of advantages. First graph based methods are invariant to input ordering of aircraft. In related work by Brittain [3] the *N*-nearest aircraft are considered. Here the order of the *N*-nearest aircraft is important since a permutation in this ordering changes the input considerably. The model will have to learn to cope with this. When using graph based methods this is not a problem. Second, graph based methods can easily deal with a different number of neighbors. This offers an advantage to considering a fixed number of aircraft. Lastly, they can easily handle a variable number of nodes (or aircraft in this context).

In this paper graph based two deep learning approaches applied and compared. The first technique is the graph convolutional (GCN) neural network layer, the second is the graph attention (GAT) neural network layer. Both approaches take the same two inputs:

1. The feature matrix $X \in \mathbb{R}^{N \times D}$ where $N$ is the number of nodes and $D$ is the number of input features.

2. The adjacency matrix $A \in \mathbb{R}^{N \times N}$ which describes the connectivity of the graph.

Here $N$ is the number of nodes (or aircraft) and $D$ in the input dimension. Both approaches output a matrix $Z \in \mathbb{R}^{N \times F}$ where $F$ is the output dimension. To understand the difference between these two approaches, consider two connected nodes, $i$ and $j$. The connection strength between these two nodes is given by $\alpha_{ij}$. The GCN layer computes $\alpha_{ij}$ by taking an average weighted by the number of connected nodes. The GAT layer implicitly learns the value for $\alpha_{ij}$ during training, such that the important node have the largest weights, etc. This is a more powerful method which comes at the expense of more parameters in the neural network layer, and thus a higher computational complexity. For a detailed and technical explanation of GCNs, GATs and their differences the reader is referred to [9], [19] or [20].

IV. PROBLEM FORMULATION

*A. Experiment*

To conduct research on the performance of DRL models on the task of air traffic controller, an experiment was designed. This experiment consists of an airspace controlled by the air traffic control model. The goal of the control model is to steer each aircraft to their desired altitude, heading and speed while avoiding crashes and conflict between pairs of aircraft that are too close to each other.

The performance of the controller is evaluated on traffic densities resembling real world traffic densities. For this, traffic densities inside the Maastricht Upper Area Control center (MUAC) are used. This is an airspace above The Netherlands, Belgium, Luxembourg and parts of Germany of approximately $260,000 km^2$. In 2018, on average, it handled 4,900 aircraft per day. MUAC handles aircraft flying at 25,000ft (7.6km) and higher. It is the third busiest upper control area in terms of flight numbers in Europe but the busiest in terms of flight hours and distance[2]. In this experiment a 24 hour period is simulated with traffic levels of an average day in 2018. The controller controls a circular airspace with a radius of 150 kilometer where randomly generated aircraft want to fly across this airspace. The number of traffic movements is proportional to the traffic movements inside the MUAC resulting in approximately 1,300 overflights per day. Furthermore, the traffic density is increased by 1.5× to measure the performance on higher than normal traffic densities.

*B. Simulator*

In this research the publicly available, python written simulator from CSU Stanislaus is used. In this simulator the goal is to guide an aircraft across the airspace, which resembles the task of an en-route controller. It does not allow for more complicated maneuvers like take-offs, landings or holdings. Other options do exist, like the BlueSky ATC Simulator Project [7]. The reason for choosing this simulator was the possibility of a fast time simulation and the ease of implementation[3].

---

[2] Numbers from the MUAC 2018 Annual Report: *https://www.eurocontrol.int/sites/default/files/2019-08/muac-annual-report-2018.pdf*

[3] Simulator available here: *https://github.com/devries/flight-control-exercise*

### 1) Initialization

In this simulator, aircraft are initialized randomly on the border of a circular airspace. Next to the initial conditions, each aircraft is given a desired heading, speed and altitude. These values are initialized uniformly from a given range.

- Initial heading: from {0, 5, …, 350, 355}°

- Initial speed: from {215, 220, …, 245, 250}m/s

- Initial altitude: from {6000, 6100, ..., 9900, 10000}m

- Desired heading: initial heading + sample from {-30, -25, …, 30} °

- Desired speed: from {215, 220, …, 245, 250}m/s

- Desired altitude: from {6000, 6100, …, 10000}m

Aircraft are initialized in pairs in such a way that given their initial state a crash will be guaranteed when no action is undertaken by the controller. The goal of the air traffic control model is to avoid crashes, and after that, conflicts.

### 2) Action space and state space

Every 5 seconds, the controller assigns each aircraft with an action. It is not possible for an aircraft to perform multiple actions in the same time step. Actions are discrete, and limited to 7 possibilities.

| | |
|---|---|
| 1. Take no action | 5. Decrease speed 5kts |
| 2. Climb 100m | 6. Turn left 5° |
| 3. Descend 100m | 7. Turn right 5° |
| 4. Increase speed 5kts | |

The state of each aircraft is defined by the 8-tuple $(x, y, z, h, s, z_{diff}, s_{diff}, h_{diff})$, where $x, y, z$ are the x, y and z coordinate of the aircraft, $h$ and $s$ are the aircraft heading and speed, and $z_{diff}$, $s_{diff}$ and $h_{diff}$ are the difference to the aircrafts desired altitude, speed and heading, respectively. For easier learning the difference between the current state and the desired state is incorporated, for example $z_{diff} = z_{des} - z$. The states of all aircraft is used as input to the air traffic control model.

### 3) Proximity to other aircraft

The proximity to other aircraft consists of two components, a vertical one and a horizontal one. A small horizontal separation between aircraft need not be a problem if the vertical separation is large enough, and vice versa. In the implementation in this research each aircraft has two cylinders around itself ($C_1$ and $C_2$), the one larger than the other. In the smaller cylinder ($C_2$) other traffic is considered too close for comfort while in the larger cylinder ($C_1$) traffic is observed, but is not found to be too close. The smaller cylinder is called the penalty area and the larger cylinder is called the detection area. This is similar to how modern traffic collision avoidance systems work [11] and is shown in figure 2 with a top view and a side view. If an aircraft enters the penalty area of another aircraft it is counted as a conflict and one of the two aircraft is set to uncontrollable: commands are no longer issued to this aircraft and only one aircraft is allowed to perform evasive actions. Results acquired from this simulator using this approach were better than without and it is also not uncommon in literature [5].

### 4) Performance metrics

The performance of the air traffic control model is given in terms of five performance metrics:

1. The number of crashes during the 24 hour simulation.

2. The percentage of potential conflicts solved. Since aircraft are initialized in pairs on a conflicting course, each pair of aircraft is counted as a potential conflict.

3. The average delay of an aircraft crossing the airspace compared to the nominal time. The nominal time is the time it would take the aircraft to cross the airspace with no other aircraft in the airspace.

4. The number of maneuvers an aircraft took compared to the nominal number of maneuvers.

5. The percentage of correct exits. A correct exit is defined as an aircraft exiting the airspace at its desired altitude, heading and speed.

The authors are aware that the last performance metric is a bit uncommon. Since we consider free airspace without waypoints, we chose to define a correct exit based on altitude, heading, and speed rather than position. Changing the implementation to take waypoints into account is considered for future work.

## V. IMPLEMENTATION

### A. Air traffic control model

The policy for the controller is implemented as a neural network. The actor and the critic share a similar architecture, but do not share weights. The actor and critic take four inputs: the aircraft states and three different adjacency matrices. The state of each aircraft is first projected through two feed-forward layers, first to 64 dimensions and then to 128 dimensions resulting in an embedding of the aircrafts state.

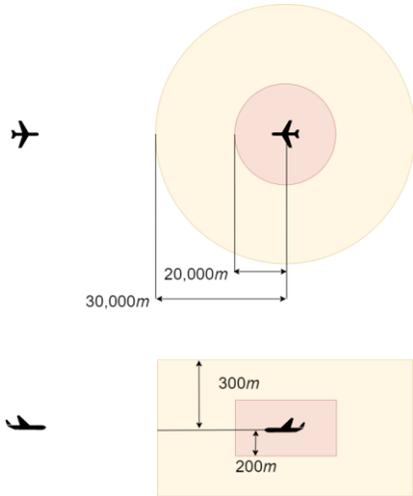

Figure 2. Visualization of the detection area and penalty area. Point of view is the aircraft on the right. The aircraft on the left is considered the intruding aircraft. The numbers in this figure are based on realistic separation requirements and finetuned for this research.

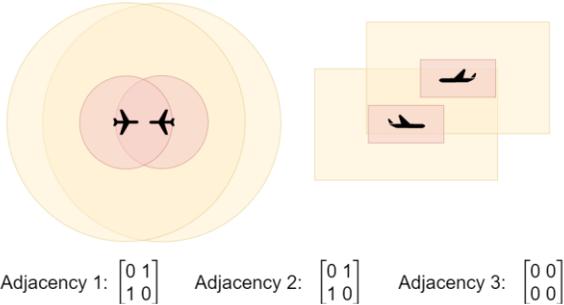

Figure 3. Situation sketch of two aircraft and their adjacency matrices.

The different adjacency matrices take the proximity to other aircraft into account in different ways, giving the model information on how far away an aircraft is. The aforementioned embedding is propagated three times through a graph convolution or graph attention layer using the three different adjacency matrices resulting in three 128 dimensional vectors. Implementing three parallel graph based layers with different adjacency matrices allows the aircraft to have multiple levels of understanding of its surrounding, providing more information if aircraft are closer.

The first adjacency matrix takes global information into account, allowing each aircraft to always have a view of its surrounding. It provides an aircraft with information on what other aircraft there are except itself. This translates into an adjacency matrix filled with non-zero elements except on the diagonal. The second adjacency matrix only takes aircraft into account inside the detection area. This is the yellow part of Figure 2. The third adjacency matrix only takes aircraft into account inside the penalty area. This is the red part of figure 2. A situation sketch of two aircraft is given in figure 3.

A skip connection is then applied to the three graph convolutional layers and the embedding of the state. These four vectors are then summed, and propagated through a feed-forward layer to 64 dimensions. This is then propagated through another linear layer, either resulting in a 7 dimensional vector if the action head is considered, or a 1 dimensional vector if the value head is considered. All feed-forward layers have ReLU activation functions, except the final layer on the action and value head, which have a SoftMax and linear activation. After summing the four 128 dimensional layers another activation is applied. A summary of the model is shown in figure 4.

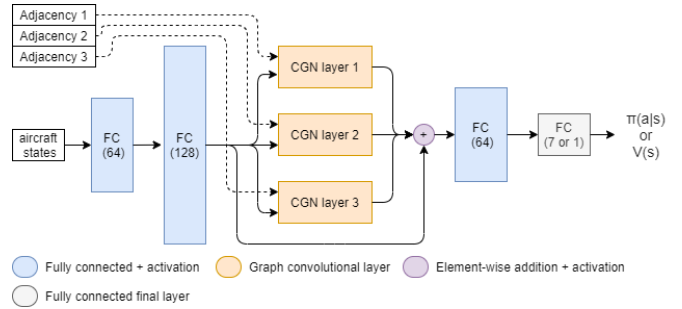

Figure 4. Visualization of the neural network architecture with shared architecture for actor and critic. The graph convolutional layers can be replaced by graph attention layers.

### B. Reward function

The reward function is an addition of the base reward function and the shaping reward function. The shaping reward function is of the form presented in equation 6. Depending on the number of neighboring aircraft, different base reward functions are used as shown in Equation 7. The potential function $\Phi(s)$ is also dependent on the state of the aircraft and its neighbors, and is given by equation 8. Inspiration for using $L_1$ norm is drawn from [14] where the negative $L_1$ norm is used as shaping function in a grid world environment. This function penalizes steps taken away from the goal, and rewards steps taken towards the goal.

$$R(s, a, s') = \begin{cases} 1 & z_{diff} \wedge s_{diff} \wedge h_{diff} = 0 \\ -1 & \text{aircraft in penalty area} \\ -100 & \text{crashed} \\ 0 & \text{otherwise or aircraft not controllable} \end{cases} \quad (7)$$

$$\Phi(s) = \begin{cases} -||s - goal||_1 & \text{no aircraft in } C_1 \text{ and } C_2 \\ \text{Equation (9)} & \text{aircraft in penalty area} \\ 0 & \text{otherwise or aircraft not controllable} \end{cases} \quad (8)$$

$$\Phi(s) = -b + |c_1 \cdot x + c_2 \cdot y| + |c_2 \cdot y - c_1 \cdot x| \quad (9)$$

Equation (9) is the form of an inverted rectangular pyramid, where $x$ and $y$ are the horizontal and vertical distance to the aircraft closest neighbor, and $b$, $c_1$ and $c_2$ are scaling constants.

When applied in (6) this function ensures a positive reward if the horizontal or vertical distance to the closest neighboring aircraft is increased.

### C. Training procedure

Training is done by simulating an episode by continuously randomly generating a pair of aircraft that are guaranteed to crash if no action is taken. Initializing this way forces aircraft to fly towards each other, and allows the learned policy to take action on this. The algorithm used for training the policy is the actor-critic algorithm with GAE described in section 3.A.1.

The maximum number of aircraft in the airspace is limited to 10. When aircraft exit the airspace they are removed. Two new aircraft are added if doing so does not cause the maximum number of aircraft to be higher than 10. The environment is simulated for 5 seconds after every action to allow clear transitions from one state to the next. The episode is terminated after 30 aircraft have been created or if a fixed number of steps is reached. Training is terminated after 5,000 episodes of training.

### D. Comparison to other work

The closest related work is the work by Brittain and Wei in [3] and [4]. This work considers a two-dimensional structured airspace where the goal is to avoid conflicts on intersecting and merging airways. Their method is not directly applicable to the free airspace considered in this paper. Because their case study differs a lot from ours and the two methods are not one on one comparable, a direct comparison could not be performed.

### VI. RESULTS

Results are obtained by training each model five times in five different training runs with different seeds. Then the experiments is performed five times for each of the five models. The median and interquartile range (IQR) of these 25 experiments are reported. The median is reported because some runs tend to produce outliers, which the median is robust to. The $1\times$ traffic density results in, on average, 11 aircraft in the airspace. The maximum number of aircraft is 25. This is already $2.5\times$ higher than during training where the maximum number of aircraft is set to 10. For the $1.5\times$ traffic density the average and maximum number of aircraft is 16 and 35.

From the results presented in table I it can be seen that the graph attention (GAT) approach is superior compared to the GCN approach in separating aircraft and avoiding conflicts. Both approaches allow for communication between aircraft, either via the convolutional mechanism or via the attention mechanism. However, the graph convolution mechanism receives the average of the aircrafts neighbors features. In a congested airspace this results in taking the average over many aircraft and information from individual aircraft can be lost, explaining the poorer performance of the GCN approach compared to the GAT approach in avoiding crashes (0 to 5) and solving conflicts (89.8% to 85.6%).

The graph attention based method is able to separate aircraft safely and efficiently under normal ($1\times$) traffic flow. The graph convolution based method cannot cope with this number of aircraft in the airspace resulting in crashes. When increasing the traffic flow to $1.5\times$ normal traffic flow both methods have a difficult time coping with the congested airspace. An analysis of the conflicts shows that conflicts happen when the number of aircraft in the airspace is higher than normal. This can be explained by the fact that the number of aircraft in the experiment is higher than seen during training (25 to 10).

An analysis of the interquartile range shows that the graph convolutional approach is the most unstable. The average delays and average number of maneuvers more than necessary fluctuate. Reasons for this are discussed below. The graph attention approach is very stable over multiple seeds even at high traffic densities. The graph convolutional approaches tend to avoid collisions by separating aircraft vertically and by changing their heading. The graph attention approach also changes the altitude of aircraft on a collision course but is less inclined to change their heading too. This explains the difference in the average delay column between the two methods. A possible explanation for this could be that the graph convolution mechanism receives the average of the aircrafts neighbors features. In a congested airspace information from individual aircraft can be lost. To cope with this, the graph convolution approach learns to prevent collisions by changing the altitude and the heading, which might be safer than just climbing. Sometimes this can result in spirals, which greatly add to the delay and number of maneuvers. This also explains the high interquartile range of the graph convolutional approach. The graph attention mechanism is able to send different kinds of messages to its neighbors and suffers less from this problem.

The high number of maneuvers needed by the graph based approaches can be explained by the way the model and the environment are implemented. Aircraft are given spatial information by three different adjacency matrices. These matrices contain information about the proximity to other aircraft. However, the exact distance to other aircraft is unknown, it only knows that there are other aircraft in $C_1$ or $C_2$. When an aircraft passes another aircraft it usually does so by changing altitude.

TABLE I. RESULTS OF THE 24 HOUR SIMULATION EXPERIMENT. THE MEDIAN AND IQR ARE REPORTED. RESULTS IN **BOLD** INDICATE THE BEST RESULTS IN THAT COLUMN.

| Method | Traffic density | Avg delay (s) | Avg # mnvr. more than necessary | # of crashes | % of potential conflicts solved | % correct exit |
|--------|------------------|---------------|-------------------------------|--------------|--------------------------------|-----------------|
| | | Lower is better | | | Higher is better | |
| GCN | 1 x | 45.1 (91) | **60.5 (27)** | 5 (5) | 85.6 (4) | 68.8 (6) |
| | 1.5 x | 58.6 (101) | **64.1 (23)** | 8 (5) | 82.9 (4) | 67.6 (6) |
| GAT | 1 x | **5.3 (2)** | 64.5 (5) | **0 (1)** | **89.8 (2)** | **75.5 (3)** |
| | 1.5 x | **5.6 (3)** | 67.7 (6) | **1 (1)** | **88.1 (1)** | **72.2 (2)** |

Consider two aircraft, *A* and *B* on a collision course and both on their desired altitude. To prevent a collision one aircraft, say aircraft *A*, will climb. It will keep climbing until aircraft *B* has exited its $C_2$. Aircraft *A* will then descend towards its desired altitude but by descending it may again enter the $C_2$ of aircraft *B*. It has learned that this should be avoided and thus climbs again. This results in oscillatory vertical movements causing the high number of maneuvers.

## VII. CONCLUSION

This paper presents a first step towards a learned free airspace autonomous air traffic control model capable of performing the task of an en-route controller. In the 24-hour simulation experiment the graph attention based model developed in this research has learned to steer aircraft to their desired altitude, heading and speed while preventing collisions. On normal traffic densities it is capable of prevent 100% of potential collisions and 89.8% of potential conflicts. However, performance deteriorates when the traffic density increases. Overall, the graph based methods used in this research proved to be a very suitable framework for this air traffic control problem and are an improvement with respect to current state of the art methods. This is because graph based methods are invariant to the ordering of aircraft and are invariant to the number of aircraft. This research is the first time that deep reinforcement learning techniques are applied on the three-dimensional, unstructured airspace, air traffic control problem. Thus, providing other researchers with a starting point for future work is an important contribution of this research. Future research could focus (among other things) on adding stochastic variables like weather, removing the oscillatory movements, adding waypoints or changing the simulator. Changing to the BlueSky simulator would make this work more easily comparable to other work.